\documentclass[12pt, letterpaper]{article}
\usepackage{graphicx} 
\usepackage{subfig}
\usepackage{tabularx}
\usepackage{float}
\usepackage[colorlinks]{hyperref}
\graphicspath{{./images/}}

\title{Simulating Malaria Detection in Laboratories using Deep Learning}
\author{Onyekachukwu R. Okonji \\ onyekaokonji@gmail.com}
\date{}

\begin{document}

\maketitle

\begin{abstract}
    Malaria is usually diagnosed by a microbiologist by examining a small sample of blood smear. Reducing mortality from malaria infection is possible if it is diagnosed early and followed with appropriate treatment. While the WHO has set audacious goals of reducing malaria incidence and mortality rates by 
    90\% in 2030 and eliminating malaria in 35 countries by that time \cite{WHO}, it still remains a difficult challenge. Computer-assisted diagnostics are on the rise these days as they can be used effectively as a primary test in the absence of or providing assistance to a physician or pathologist. The purpose of this paper is to describe an approach to detecting, localizing and counting parasitic cells in blood sample images towards easing the burden on healthcare workers.    
\end{abstract}

\section{Introduction}
Malaria is one of the top 10 diseases which claims the highest mortality rates globally. According to the 2022 factsheet \cite{WHO} released by the WHO, there were 241 million cases of malaria infection globally with 619,000 deaths globally in 2021. It is estimated that the numbers will continue to rise leading to a greater increase in mortality rates. According to the WHO, early diagnosis and treatment reduces disease rates, deaths and contributes to reducing transmission \cite{WHO} but this still proves difficult especially in regions of the world particularly Sub-Saharan Africa with the largest proportion of malaria cases and challenges with detection ranging from inadequate testing facilities to overburdening of the few available. To contribute towards tackling this challenge, we train a machine learning model which analyzes thousands of images that were collected in rural areas and processed in order to automatically detect Malaria simulating laboratory settings towards reducing the workload of physicians and pathologists. 

\section{Related work}
Improving the detection of malaria parasites in blood sample images has been the subject of a number of research efforts. Majority of which involve the use of convolutional neural networks for their ability to extract distinctive features in images and then make accurate predictions. \cite{RL1} involves the use of a combination of blob detection for recognizing regions in images that differ from the whole image with respect to color and brightness for example, followed by feature extraction using GoogLeNet architecture, the output of which is passed to a SVM layer for classification. \cite{RL2} proposes a mobile application for the automatic detection of malaria, especially in low-resource communities. Their technique involves fixing a mobile phone to a microscope, taking a picture of the slides, which are then sent to a database for each patient and inference performed. Their technique also allows for automatic updating of the machine learning models between phones of health workers using bluetooth connection. \cite{RL3} proposes a novel bin approach to image classification, the technique focuses on improving the quality of the features detected in an input image across each channel of the image, with the goal of improving classification with the feature vectors obtained were passed to SVM, RNN and KNN layers for final image classification.

While all these are great tasks, we notice that most of them do not completely simulate the real-world setting as they mostly stop at classification of the blood sample images without goint further to localize and count the number of detected cells. This advanced step is what this paper aims to achieve.

\section{Methodology}
Our approach is one which aims to closely replicate the steps and motivations being done in labs everyday. As such the steps include:

\begin{itemize}
    \item Image processing of input images: Gaussian Blur and Edge detection for the purpose of removing noise in images and to highlight visible features which may aid accurate classification.
    \item Training of a convolutional neural network on the input images with the aim of achieving a model with high accuracy, precision and recall.
    \item Applying a sliding window over the images classified as containing malaria parasites with the aim of localizing and counting the number of parasitic cells in the image.
    \item Inference using mobile-phone based/offline techniques. The offline technique is aimed at displaying the potential of use of the model in low resource and underprivileged societies.
\end{itemize}

\section{Image processing techniques}
Image analysis or processing is the extraction of meaningful information from images; mainly from digital images by means of digital image processing techniques. Dataset was obtained from Kaggle \cite{dataset}. As we need to feed images of similar spatial dimensions, we resize the images into 75px by 75px using OpenCV’s cv2.resize() object \cite{RESIZE}. Upon resizing the images, we sought to perform edge detection on the images with the aim of highlighting features which may help the model make a more accurate prediction. For edge detection we first applied a Gaussian filter \cite{GAUSSIAN} using a 7 x 7 kernel to reduce any noise in the image which could negatively impact classification, after which we apply the canny edge detector \cite{Canny} from OpenCV which uses a multi-stage algorithm to detect a wide range of edges in images to make image edges sharper and image smoother with threshold values of 80px and 160px The result can be seen using a sample image in Figure \ref{Fig. 1} below.

\begin{figure}[H]
    \centering
    \subfloat[original image]{\includegraphics[width=1.6in]{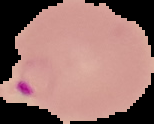}}
    \hfill
    \subfloat[processed image]{\includegraphics[width=1.6in]{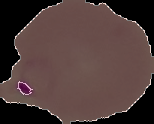}}
    \caption{Left is the original image before processing. Right is the image post-processing}
    \label{Fig. 1}
\end{figure}

\section{Model development choices and training}
Taking into consideration the No Free Lunch theorem \cite{NFLT} which when applied to Machine Learning summarily states that “no machine learning algorithm is better than the other”, we train several models on our dataset with the hope of finding the best model that fits to the distributions of data in the real world. As such, we train on 9 different state-of-the-art algorithms in visual recognition tasks, via the pre-trained approach using Keras, \cite{appli} the results of which can be found in Table 1 below:

\begin{enumerate}
    \item \textbf{VGG} \cite{VGG}: Due to the success of convolutional nets in image recognition tasks, a number of techniques have been used to improve their accuracy, ranging from using smaller receptive windows and strides to training and testing the networks over whole images and multiple scales. The authors behind VGG thought to take a different approach which involved working with the depth of the model - increasing model depth. Taking a detour from prior state-of-the-art conv net architectures which used large (5 x 5 to 7 x 7) receptive fields with smaller strides, VGG used a 3 x 3 and 1 x 1 receptive field with smaller strides (1 x 1) this provided the advantage of decreased number of parameters, and with the use of more non-linear rectified activations, this made the model’s decision function more discriminative, improving its generalizability. The introduction and success of VGG architecture birthed a new wave of research into deep neural networks. For this task, we train both VGG16 and VGG19 architectures on our data.
    
    \item \textbf{Inception} \cite{Inception}: The Inception architecture aims at approximating the optimal sparse structure similar to biological systems through the use of readily available dense building blocks. It consists of inception modules which are aimed at finding the optimal local construction and repeating it spatially. The modules consist of filters of size 1 x 1, 3 x 3 and 5 x 5 and these modules are stacked on top of each other forming the architecture. The architecture comes with a slight increase in computation cost with a greater increase in accuracy in visual recognition tasks. For this task, we train both InceptionNetv3 and InceptionNet-ResNetV2 architectures on our data.
    
    \begin{figure}[h]
        \centering
        \includegraphics[width=2.59in]{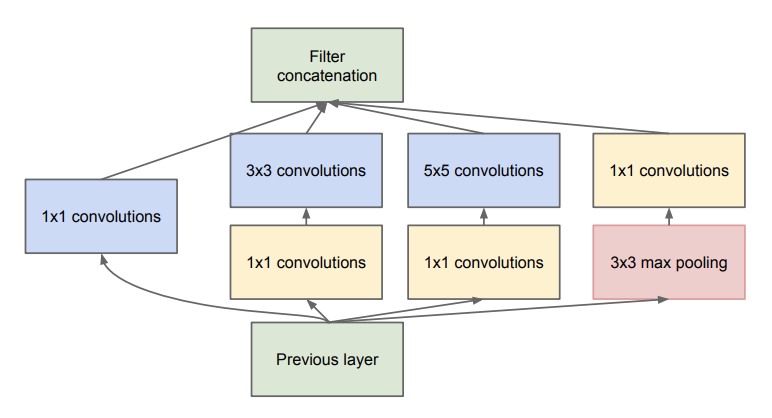}
        \caption{inception module}
        \label{Fig. 2}
    \end{figure}
    
    \item \textbf{Xception} \cite{Xception}: Taking inspiration from the work done on the Inception class of architectures, Xception aims at achieving the same purpose via a different means. Unlike the inception modules in the Inception framework, Xception uses depthwise separable convolutions which are a combination of depthwise convolutions and pointwise convolutions, stacked upon each other to make the framework. It also uses residual connections between each stack which helps with speed and model performance. When compared with the InceptionV3 member of the Inception family, it offered nearly the same number of parameters but with a slight increase in accuracy.
    
    \begin{figure}[h]
        \centering
        \includegraphics[width=2.48in]{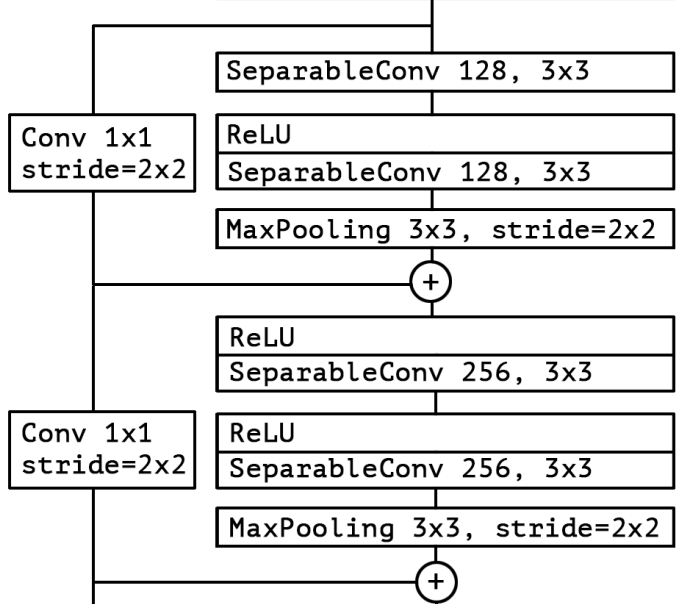}
        \caption{ two blocks of depthwise separable convolutions with residual connections which form the central part of the Xception architecture}
        \label{Fig. 3}
    \end{figure}

    \item \textbf{ResNets} \cite{ResNet}: Because deep networks were proven to perform better at visual recognition tasks and competitions, new models were being introduced with increasing depth in the layers of the model. This came with a few challenges, first being the issue of exploding and vanishing gradients, the latter occuring as a result of the gradients getting diminished or greatly reduced as they make a backward pass through the network which by extension makes training difficult. It was observed that the deeper a neural net is, the higher the chances the model’s accuracy saturated and this could not be considered an overfitting problem where deeper models can be a solution. Thus, ResNets aimed to tackle this challenge via the introduction of identity mappings of the output feature maps of each layer being passed via element-wise addition to the next layer. During experiments, it was observed that this identity mappings neither increase model parameters nor increase error rates. On the contrary, they reduced the error rates and made the models more generalizable to validation data during training. Of the variants of ResNet architectures being used, ResNet-152 (the number representing the number of layers), performed best and has lower complexity to previous deep nets notably VGG-16/19 and also enjoys high accuracy. For this task, we train ResNet-152 architecture on our data.
    
    \begin{figure}[h]
        \centering
        \includegraphics[width=2.68in]{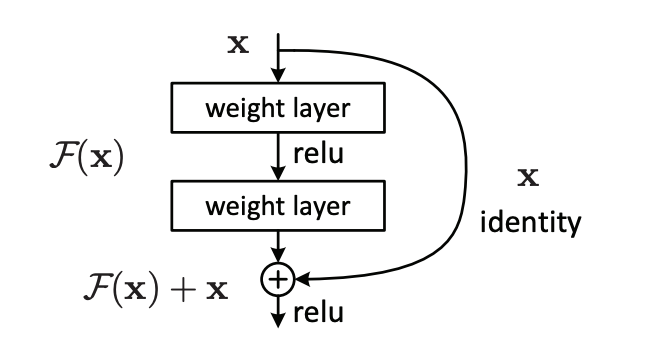}
        \caption{A residual connection between a layer and successive layers}
        \label{Fig. 4}
    \end{figure}

    \item \textbf{DenseNets} \cite{DenseNet}: Introduced in 2018, the DenseNet architecture contributes to techniques aimed at information preservation in deep neural networks. Coming at the back of successes (and challenges) with preceding architectures, notably ResNets and InceptionNets, DenseNets take advantage of feature re-use as a way to preserve information flow. What this means is, as opposed to combining features of each layer with those of preceding layers via addition (as is the case in ResNets), DenseNets use concatenation technique, such that the output feature map of each layer is concatenated with those of preceding layers, the output of which is sent to subsequent layers. Compared to ResNets, the number of parameters is considerably smaller and it has the added advantage of improved flow of information and gradients through the network, thus making it easier to train. Compared to InceptionNets which also concatenate feature maps, making the network wider, the DenseNets are simpler and more efficient. For this task, we train DenseNet-121 architecture on our data.
    
    \begin{figure}[h]
        \centering
        \includegraphics[width=2.83in]{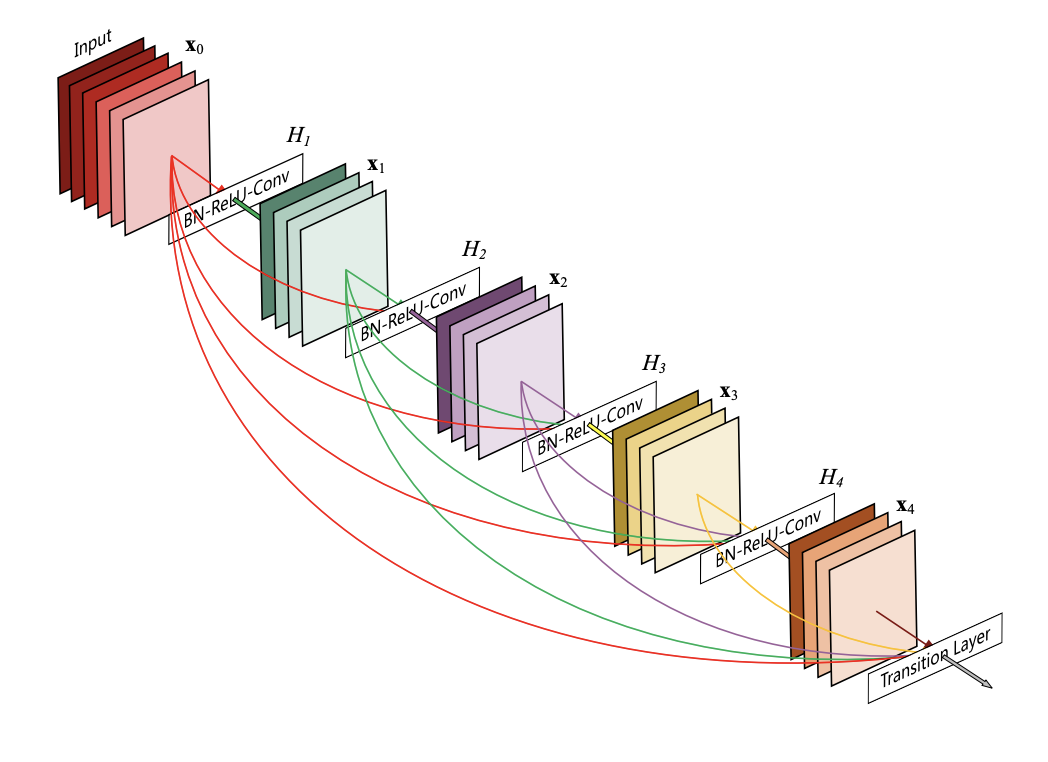}
        \caption{DenseNet architecture}
        \label{Fig. 5}
    \end{figure}

    \item \textbf{MobileNet} \cite{MobileNet}: the idea behind MobileNet class of architectures is to build models of considerably reduced number of parameters and computational cost with the added benefit of reduced latency with a trade-off in accuracy. It works via a technique called depthwise separable convolution which is a combination of depthwise and pointwise convolutions. Depthwise convolutions involve applying a 3 x 3 filter to every channel of the input. The output of which is then combined via a 1 x 1 convolution which serves as the pointwise convolution. The use of depthwise convolutions results in significantly reduced cost. MobileNets have 2 in-built hyperparameters which can be adjusted to further reduce computational costs and size. The first being the width multiplier $\alpha$ which within the range (0,1) can be applied to the input and output channels of the framework further reducing it’s computational costs, the second hyperparameter being the resolution multiplier $\rho$ which serves to reduce the resolution of the input image, $\rho$ exists within the range (0,1) and the default values for $\alpha$ and $\rho$ is 1.For this task, we train MobileNetV2 architecture on our data.
    
    \item \textbf{EfficientNet} \cite{EfficientNet}: focuses on scaling up convnets where contrary to previous architectures which focus on one of the following network dimensions: width, depth and input resolution, EfficientNet aims to uniformly scale all dimensions of a baseline model using a set of fixed scaling coefficients. The intuition being that though scaling in one dimension (commonly depth) will improve accuracy, the accuracy eventually gets diminished as scaling is increased so it was imperative to implement a compound scaling of all 3 dimensions to get better accuracy and efficiency. The compound coefficient $\phi$ is used for this scaling. For this task, we train EfficientNetV2S architecture on our data.
\end{enumerate}

Hyperparameters used during training include:

\begin{itemize}
    \item Number of epochs: 30
    \item Learning rate: 0.0001
    \item Callback: custom callback for stopping training if validation accuracy exceeds 98$\%$ in addition to reduction in learning rate when training accuracy plateaus
    \item Optimizer: Adam
\end{itemize}

\begin{table}[H]
\begin{tabularx}{1.0\textwidth}{ | >{\centering\arraybackslash}X 
    | >{\centering\arraybackslash}X 
    | >{\centering\arraybackslash}X
    | >{\centering\arraybackslash}X
    | >{\centering\arraybackslash}X | }
    \hline
    Model architecture & Total number of parameters & Validation accuracy ($\%$) & Validation precision ($\%$) & Validation recall ($\%$)\\
    \hline
    MobileNetV2 & 9.2M & 93.6 & 92.3 & 92.6 \\
    \hline
    VGG19 & 20.3M & 90.5 & 86.5 & 93.9 \\
    \hline
    ResNet-152 & 60.7M & 78.01 & 77.68 & 77.27 \\
    \hline
    InceptionV3 & 24.4M & 88.62 & 90.84 & 85.50 \\
    \hline
    Inception-ResNetV2 & 55.2M & 91.85 & 87.94 & 94.47 \\
    \hline
    EfficientNetV2S & 32M & 79.41 & 76.09 & 84.13 \\
    \hline
    Xception & 39.5M & 92.40 & 94.43 & 87.37 \\
    \hline
    \textbf{DenseNet-121} & \textbf{11.2M} & \textbf{93.80} & \textbf{94.10} & \textbf{92.53} \\
    \hline
    VGG16 & 15M & 90.98 & 92.24 & 89.30 \\
    \hline
\end{tabularx}
\caption{Results from training 9 models on the images}
\label{table: 1}
\end{table}

\section{Parasite localization}
Taking cognizance of the fact that in the laboratory setting, after the pathologist has concluded that a given blood sample contains the malaria parasites, they then begin to count the number of cells present, we sought to expand the scope of the task by counting the number of parasitic cells present in a given image and then localize them for easier viewing through a computer. The purpose of this is to automate the manual process of counting the cells which is error-prone and time consuming. To achieve this, we used the age-old sliding window approach for object detection. To do this, we take the positively-classified images, generating pyramids of different sizes of the image, the aim being to train the sliding window to detect objects of varying sizes. The sliding window goes through all rows of the image in question, detecting the presence of objects (parasitic cells) and then drawing bounding boxes around each detected cell.

Using the sliding window approach can be quite hard-coded and sub-optimal across different settings hence this is a likely area for future research - other techniques at transforming image classification tasks to object detection or localization tasks while retaining both capabilities.

\begin{figure}[H]
    \centering
    \includegraphics[height=2.59in]{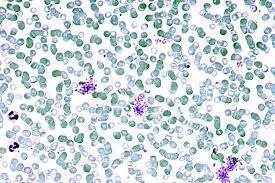}
    \caption{Input image for parasite localization and Output image after applying sliding window}
    \label{Fig. 6}
\end{figure}

\section{Tackling model explainability}
Despite the fact that CNNs have shown significant and reliable performance in computer vision tasks particularly image classification, there have been concerns about the lack of understanding of how a CNN model makes its decisions. This led to CNN models being called “black box” implying the difficulty in knowing what features contributed to the model’s output prediction. This is in contrast to other Machine Learning techniques and algorithms like Linear Regression and Decision Trees which it is quite easy to understand their “thought-process” and so are called “white box”. This lack of explainability remains a huge factor to its adoption in workflow optimization. Explainable AI is a term to describe methods aimed at monitoring the performance of Machine Learning algorithms and improving trust in the ML process. An example use case in a healthcare setting is one where a physician would like to know what parts of an image the model focuses on in order to make its predictions. To the physician, it is not enough to see a prediction though accurate and then decide to go with it, it is important to know for sure that the model actually focused on the right aspects of the input and so wasn’t merely making correct guesses. For a Machine Learning Engineer, explainability helps the engineer to know what aspects of the input contributed significantly to the decision-making potential of the model. A number of techniques have been used over the years at improving the explainability of models. Some of which include, Shapely Adaptive exPlanations \cite{SHAP} (SHAP) which is a technique to measure how each features of the input data contribute to the prediction of the model, Local Interpretable Model Explanations (LIME) \cite{LIME}, which have the same goal as SHAP but do so by approximating a black box model with a white box, explainable mode for each prediction, Saliency Maps \cite{Saliency} which essentially measures the unique features of a given image and Class Activation Maps (CAM), one of which is Grad-CAM \cite{GradCAM} which uses the gradients being propagated to the final convolutional layer of a model to obtain coarse representations of the relevant areas in the image influencing the model’s prediction. In computer vision projects, CAMs are commonly used as the best method of improving model explainability. 

CAM was first introduced in the MIT paper; Learning Deep Features for Discriminative Localization \cite{cam}. For any given label or classes to be predicted, Class Activation Maps help to single out the discriminative part of the image which helped the model make its predictions. CAMs work by applying a Global Average Pooling layer to the last convolutional layer of a given CNN, with the aim being the aggregation of the feature maps necessary for classification. Average pooling layers are used over the common Max pooling layers as the goal here is to have a good representation of all discriminative regions in the particular layer, a max pool will only pick the most discriminative regions. 

In this paper, we make use of a Python package which helps to accurately display class activation mappings of a model on any given image. Keract \cite{keract} implements the Grad-CAM approach to model explainability by fetching and generating heatmaps of the activations of each layer of a neural network. By doing so, we are able to view the activations of all layers of a model as it receives an input image. What this means is, we are able to have a good idea of what parts of an image activated the neurons of the model in order to make predictions.

\begin{figure}[H]
    \centering
    \includegraphics[height=4in]{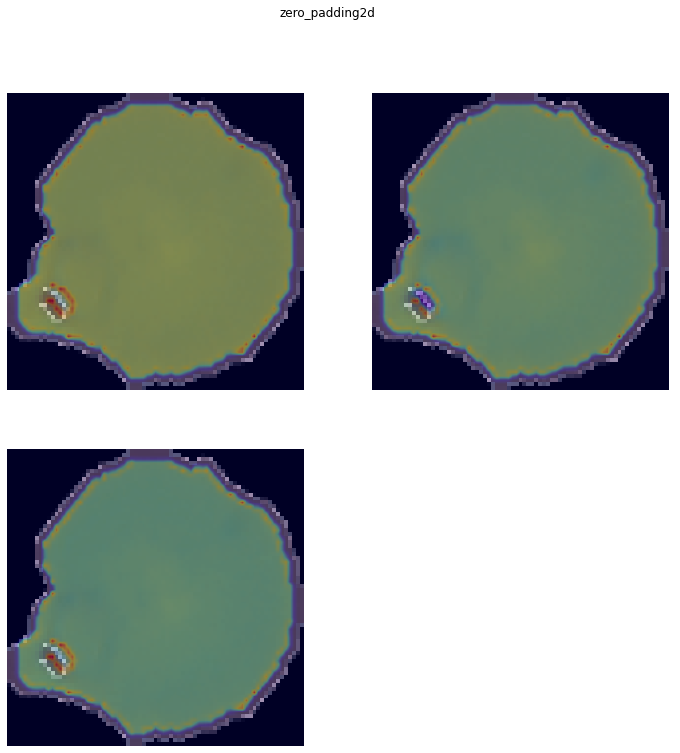}
    \caption{Heatmap showing the activations of the first layer of the DenseNet-121 trained model}
    \label{Fig. 7}
\end{figure}

\section{Deployment strategies}
\begin{figure}[H]
    \centering
    \includegraphics[height=2in]{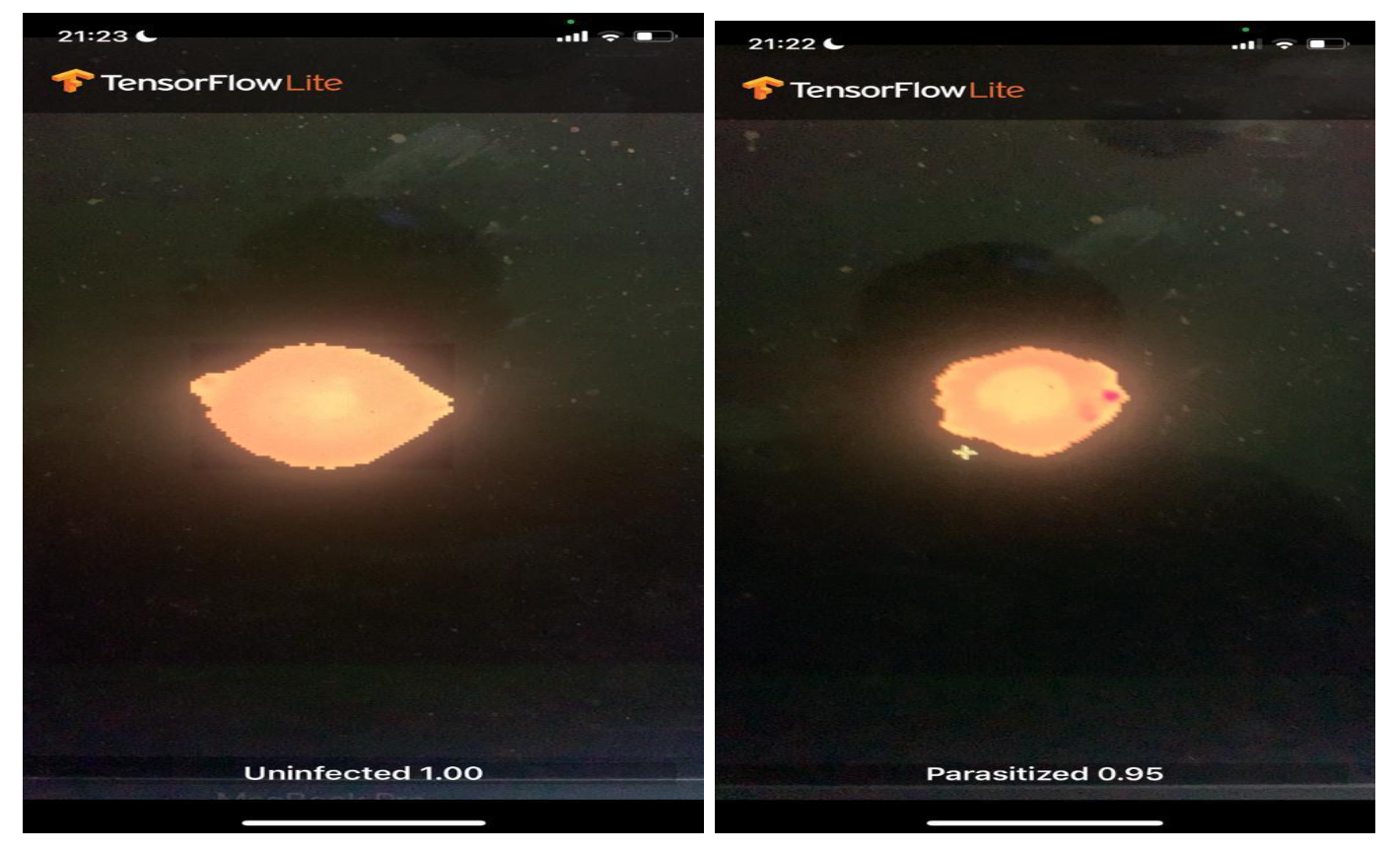}
    \caption{Mobile deployment via Tensorflow Lite and SwiftiOS}
    \label{Fig. 8}
\end{figure}

Because we sought to ensure that this model doesn’t just sit as research work and is actually useful in a clinical setting, we decided to deploy and test the model in an offline scenario using a mobile app. The reason behind the mobile app is to make the model accessible to healthcare workers in low-resource communities where access to a computer system and internet might be a challenge. Having trained all 9 models mentioned in Table \ref{table: 1} above, we selected the best perfoming model (DenseNet-121), conducted CPU inference testing to have an idea of its latency and accuracy. Satisfied with the results we got, we thought the next step will be to make the model mobile-friendly. To achieve this, we started by implementing a post-training quantization approach on the model with the aim of reducing its size at the expense of minute loss in accuracy but with the benefit of faster latency. Post-training quantization \cite{quantization} belongs to a list of techniques aimed at reducing the computational workload required to use deep learning models in production. Essentially, what it involves is a conversion of the model's weights originally of float32 types used in CNN models to 8-bit integers with a compromise in accuracy but benefit of reduced computational cost and efficiency on CPU hardware. Having done this, the next phase is to convert the quantized model to a format compatible with mobile and other edge devices, to accomplish this, we use the TensorflowLite converter \cite{tflite} from Tensorflow which takes a Tensorflow model and converts to a FlatBuffer file which can then be deployed on edge devices. Table \ref{table: 2} below highlights the drop in size of the model as we convert to the flatbuffer format and when we quantize the model before conversion while Figure \ref{Fig. 8} shows the model built on an iOS device.

\begin{table}[h]
    \begin{tabularx}{1.0\textwidth}{ | >{\centering\arraybackslash}X 
        | >{\centering\arraybackslash}X 
        | >{\centering\arraybackslash}X
        | >{\centering\arraybackslash}X
        | >{\centering\arraybackslash}X | }
        \hline
        Parameter & Original Model & TF-Lite model without Post-training quantization & TF-Lite model with Post-training quantization\\
        \hline
        Size & 76MB & 43MB & 11MB \\
        \hline
    \end{tabularx}
\caption{Noticeable is the significant reduction in the size of the model.}
\label{table: 2}
\end{table}

\section{Conclusion}
By this project, we have been able to demonstrate the possible application of convolutional neural nets in the classification of blood sample images for the presence of malaria parasites and further localizing the detected parasitic cells in the image for better workflow optimization. To make the model useful in rural clinical settings, we have also deployed a mobile app which performs inference at fast speeds using CPU memory available in mobile and edge devices making it available to healthcare workers in rural settlements without access to internet and immense computational resources.

\section*{Acknowledgment}
Sincere gratitude goes to my mentor Mr Manny Ko whose input and guidance made this possible. Can't thank you enough for always pushing me to exceed my limits.

\bibliographystyle{IEEEtran}
\bibliography{refs}
\end{document}